\documentclass{article} 
\usepackage{nips13submit_e,times}
\usepackage{hyperref}
\usepackage{url}
\usepackage{algorithmic}

\usepackage{algorithm}
\include{entete}

\newtheorem{definition}{Definition}

\newtheorem{theorem}{Theorem}
\newtheorem{corollary}[theorem]{Corollary}

\title{Domain Adaptation of Majority Votes\\via Perturbed Variation-based Label Transfer}

\author{
Emilie Morvant\thanks{This work was in parts funded by the European Research
Council under the European Unions Seventh Framework
Programme (FP7/2007-2013)/ERC grant agreement no 308036.}\\
Institute of Science and Technology Austria\\
 Klosterneuburg, 3400 Austria\\
\texttt{emorvant@ist.ac.at}
}

\nipsfinalcopy

\begin{document}

\maketitle

\begin{abstract}
We tackle the  PAC-Bayesian Domain Adaptation (DA) problem \cite{PBDA}.
This arrives when one desires to learn, from a source distribution, a good weighted majority vote (over a set of classifiers) on a different target distribution. 
In this context, the disagreement between classifiers is known crucial to control.
In \mbox{non-DA} supervised setting, a theoretical bound -- the C-bound \cite{Lacasse07} -- involves this disagreement and leads to a majority vote learning algorithm: MinCq \cite{MinCq}.
In this work, we extend MinCq to DA by taking advantage of an elegant divergence between distribution called the Perturbed Varation (PV) \cite{PV}.
Firstly, justified by a new formulation of the C-bound, we provide to MinCq a target sample labeled thanks to a PV-based self-labeling focused on regions where the source and target marginal distributions are closer.
Secondly, we propose an original process for tuning the hyperparameters. 
Our framework  shows very promising results on a toy problem.
\end{abstract}
\section{Introduction}
Nowadays, due to the expansion of Internet a large amount of data is available. Then, an important issue in Machine Learning is to develop methods able to transfer knowledge from different information sources or tasks, which is known as Transfer Learning (see \cite{pan2010} for a survey). In this work, we tackle the hard \cite{BenDavid12} problem of unsupervised Domain Adaptation (DA), which arises when we want to learn from a distribution -- the source domain -- a well performing model on a different distribution -- the target domain -- for which one has an unlabeled sample. Consider, for instance, the common problem of spam filtering, in which one task consists in adapting a model from one user to a new one. One popular solution is to take advantage of a divergence between the domains, with the
intuition that we want to minimize the divergence while preserving good performance on the source data \cite{BenDavid-NIPS07,MansourMR08,PBDA}. Some classical divergences involve  the disagreement between classifiers, which appears crucial to control. Another divergence, the Perturbed Variation (PV) \cite{PV}, is based on this principle: Two samples are similar if every target instance is close to a source instance. 
In this work, we focus on the PAC-Bayesian DA setting introduced in \cite{PBDA} for learning a good target weighted majority vote over a set of classifiers (or voters). A key point is that the divergence used, which takes
into account the expectation of the disagreement between pairs of voters, is justified by a recent tight bound on the risk of the majority vote: the C-bound \cite{Lacasse07}. This C-bound leads to an elegant and well performing algorithm for supervised classification, called MinCq \cite{MinCq}. Our contribution consists in extending MinCq to the DA scenario, thanks to a label transfer from the source domain to the target one. First, we propose in section \ref{sec:newcbound} a new version of the C-bound suitable for every label transfer defined by a label function. Then, we design in section \ref{sec:transfer} such a function thanks to the empirical PV.
Concretely, our PV-based label transfer focuses on the regions where the source and target marginals are closer, and labels the (unlabeled) target sample only in these regions. Afterwards, we provide to MinCq this auto-labeled target sample. We also make use of the PV to define an original hyperparameters validation.
Finally, we show empirically in section \ref{sec:expe} that our approach implies good and promising results on a toy problem, better than a nearest neighboorhood-based transfer.

\section{Notations and Background}
Throughout this paper, we consider the PAC-Bayesian DA setting described in \cite{PBDA}
for 
classification tasks where  $X\!\in\!\R^d$ is the  input space of dimension $d$ and $Y\!=\!\{-1,+1\}$  is the label set. The source domain $\PS$ and the target domain $\PT$ are two different distributions over $X\!\times\! Y$. $\DS$ and $\DT$ are the respective marginal distributions over $X$.  
In the PAC-Bayesian theory,  introduced in \cite{Mcallester99b}, given a set of classifiers (that we called voters) $\Hcal$ from $X$ to $\R$ and  given a prior distribution $\prior$ of support $\Hcal$,  the learner aims at finding
a posterior distribution $\posterior$ leading to a $\posterior$-weighted majority vote $B_\posterior$ over $\Hcal$ with good generalization guarantees. $\BQ$ is defined as follows.
\begin{definition}
Let $\Hcal$ be a set of voters  from $X$ to $\R$. Let $\posterior$ be a distribution over $\Hcal$. The $\posterior$-weighted majority vote $\BQ$ (sometimes called the Bayes classifier) is,\\
\centerline{$\displaystyle 
\forall \xbf\in X,\ B_\posterior(\xbf) = \sign\left[\esp{h\sim \posterior} h(\xbf)\right].
$}\\[1mm]
The true risk  of $\BQ$ on a domain $P$ is,
$\RP(\BQ)\! =\! \tfrac{1}{2} \big(1-\espdevant{(\xbf,y)\sim P} y\BQ(\xbf) \big).$ \\[-4.5mm]
\end{definition}
Usual PAC-Bayesian generalization guarantees ({\it e.g.} \cite{Mcallester03,Seeger02,Langford2005,catoni2007pac,germain2009pac}) bound the risk of the stochastic Gibbs classifier $\GQ$, which labels an example $\xbf$ by first drawing a voter $h$ from $\Hcal$ according to $\posterior$,  then returns $\sign[h(\xbf)]$. The risk of $\GQ$ corresponds to the expectation of the risks:\\[1mm] 
 \centerline{$\RP(\GQ) = \esp{h\sim \posterior} \RP(h) =  \tfrac{1}{2} \big(1-\esp{h\sim \posterior} \esp{(\xbf,y)\sim P}  y h(\xbf) \big).$}
It is then easy to relate $\BQ$ and $\GQ$ by: $\RP(\BQ)\leq 2\RP(\GQ).$\\
In that light, the authors of \cite{PBDA} have done a PAC-Bayesian analysis of DA. 
Their main result is stated in the following theorem.
\begin{theorem}[\cite{PBDA}]
\label{theo:pbda}
Let $\Hcal$ be a set of voters. For every distribution $\posterior$ over $\Hcal$, we have,\\[1mm]
\centerline{$
\RPT(\GQ)\ \leq\  \RPS(\GQ)\  +\  \des(\DS,\DT)\  +\  \lambda_\posterior,$}\\[1mm]
where $\lambda_\posterior$ is a term related to the true labeling on the two domains\footnote{Since one usually omits this term in algorithms, we do not develop it. More details could be found in \cite{PBDA}.}, and\\ $ \des(\DS,\DT)\! =\! \Big|\esp{(h,h')\sim\posterior^2}\!\!\Big(\esp{\xbf_t\sim\DT}\!h(\xbf_t) h'(\xbf_t) -\!\! \esp{\xbf_s\sim\DS}\!h(\xbf_s)h'(\xbf_s) \Big)\!\Big|$ is the domain disagreement.
\end{theorem}
This bound reflects the philosophy in DA: It is well known \cite{BenDavid-NIPS07} that a good adaptation may be possible if the divergence between the domains is small while achieving good performance on the source domain.
The point which calls our attention in this result is the definition of the domain disagreement, $\des(\DS,\DT)$, 
directly related to the disagreement between pairs of voters, 
 and justified by the definition of the following 
theoretical bound called the C-bound \cite{MinCq,Lacasse07}. 
\begin{theorem}[The C-bound as expressed in \cite{MinCq}]
For all distribution $\posterior$ over  $\Hcal$, for all domain $\PS$ over $X\times Y$ of marginal (over $X$) $\DS$, if  $\esp{h\sim \posterior}\esp{(\xbf_s,y_s)\sim \PS} y_s h(\xbf_s)> 0$, then,\\[-1mm]
\centerline{$\displaystyle \RPS(\BQ) \leq 1 - \frac{\left(\esp{h\sim \posterior}\esp{(\xbf_s,y_s)\sim \PS} y_s h(\xbf_s)\right)^2 }{\esp{(h,h')\sim \posterior^2}\esp{\xbf_s\sim \DS} h(\xbf_s)h'(\xbf_s)}.$}\\[-2mm]
\end{theorem}
Since we can remark the C-bound's denominator is also related to the disagreement between pairs of voters, we propose, in the next section, a new formulation suited for DA.
Before, we recall the supervised classification algorithm MinCq \cite{MinCq} which ensues from the C-bound (and described in Algo.~\ref{mincq}).
Concretely, MinCq learns a performing majority vote by optimizing the empirical counterpart of the C-bound: It minimizes the denominator, {\it i.e.} the disagreement  (Eq. \eqref{eq:objective}), given a fixed numerator {\it i.e.} a fixed margin for the majority vote  (Eq. \eqref{eq:mincq_constraint1}), under a particular regularization (Eq. \eqref{eq:mincq_constraint2}).\footnote{For more technical details on MinCq, please refers to \cite{MinCq}.}
Note that its consistency is  justified by a PAC-Bayesian generalization bound.\\
Since  the C-bound, and thus MinCq, focus on the disagreement between  voters, which is crucial to control in DA \cite{BenDavid-NIPS07,MansourMR08,PBDA}, we propose to make use of the C-bound and MinCq in a DA perspective. 

\begin{minipage}[t]{0.5\columnwidth}
\vspace*{-1cm}
\footnotesize \begin{algorithm}[H]\footnotesize
\begin{algorithmic}
\INPUT{A sample $S=\{(\xbf_i,y_i)\}_{i=1}^{|S|}$,  a set  of voters $\Hcal$, a desired margin $\mu>0$}
\OUTPUT{$\displaystyle \BQ(\cdot)\!=\!\sign\left[ \mbox{\tiny$\displaystyle \sum_{j=1}^{|\Hcal|}$} \left(2\posterior_j\! -\!\tfrac{1}{|\Hcal|}\right)h_j(\cdot)\right]$}
\begin{align}
\label{eq:objective}
\!\!\!\!\textbf{Solve}\ \ &\argmin{\POSTERIOR}\ \mathbf{\POSTERIOR}^T{\bf M}{\bf \POSTERIOR-A}^T{\bf \POSTERIOR},\\[-3mm]
\label{eq:mincq_constraint1} \textbf{s.t.}\ \ & \displaystyle \mathbf{m}^T{\POSTERIOR}=\frac{\mu}{2}+\frac{1}{2|S||\Hcal|}\sum_{j=1}^{|\Hcal|}
\sum_{i=1}^{|S|}y_ih_j(\xbf_i),\\[-1mm]
\label{eq:mincq_constraint2}&\displaystyle \forall j \in \{1,\ldots,|\Hcal|\},\quad 0 \leq \POSTERIOR_j \leq \tfrac{1}{|\Hcal|},
\end{align}
where  $\POSTERIOR\! =\! (\posterior_1,\ldots,\posterior_{|\Hcal|})^{T}$ is a vector of weights,\\
$\Mbf$ is the $|\Hcal|\!\times\! |\Hcal|$ matrix formed by {\tiny $\displaystyle \sum_{i=1}^{|S|}$}$\!\frac{h_j(\xbf_i)h_{j'}(\xbf_i)}{|S|}$ 
for $(j,j')\!\in\!\{1,\ldots,|\Hcal|\}^2$, 
and:\\
\centerline{\footnotesize $\displaystyle \mbf=\left(\!\tfrac{1}{|S|}\!\mbox{\tiny$\displaystyle\sum_{i=1}^{|S|}$}y_ih_1(\xbf_i),\ldots,\!\tfrac{1}{|S|}\!\mbox{\tiny$\displaystyle\sum_{i=1}^{|S|}$}y_ih_{|\Hcal|}(\xbf_i)\!\right)^{\!T}$}
 \centerline{\footnotesize$\displaystyle \Abf\! =\!\left(
\mbox{\tiny$\displaystyle\sum_{j=1}^{|\Hcal|}\!\sum_{i=1}^{|S|}\!$}\frac{h_1(\xbf_i)h_{j}(\xbf_i)}{|\Hcal||S|}\!,\ldots,\!  \mbox{\tiny$\displaystyle\sum_{j=1}^{|\Hcal|}\! \sum_{i=1}^{|S|}\!$} \frac{h_{|\Hcal|}(\xbf_i)h_{j}(\xbf_i)}{|\Hcal||S|}\right)^{\!T}$}
\end{algorithmic}
\caption{\footnotesize MinCq($S,\Hcal,\mu$) 
\label{mincq}}
\end{algorithm}
\end{minipage}\hfil
\begin{minipage}[t]{0.46\columnwidth}\vspace*{-1cm}
\begin{algorithm}[H]\footnotesize
\begin{algorithmic}
\INPUT{ $S=\{\xbf_s\}_{s=1}^{|S|}$ and $T=\{\xbf_t\}_{t=1}^{|T|}$ are unlabeled samples, $\epsilon > 0$, a distance $d$}
\OUTPUT{$\widehat{PV}(S,T)$}
\STATE 1. $G\leftarrow \big(V\!=\!(A,B),E\big)$, where $A=\{\xbf_s\!\in\! S\}$\\$\ \ \ \ $and $B=\{\xbf_t\!\in\! T\}$,  $e_{st}\in E$ if $d(\xbf_s,\xbf_t)\leq \epsilon$
\STATE 2. $M_{ST}$ $\leftarrow$  Maximum matching on $G$
\STATE 3. $S_u\leftarrow$   number of unmatched vertices in $S$\\
 $\ \ \ \ T_u\leftarrow$   number of unmatched vertices in $T$
\STATE 4. Return $\widehat{PV}(S,T)=\tfrac{1}{2}\left(\frac{S_u}{|S|}+\frac{T_u}{|T|}\right)$
\end{algorithmic}
\caption{\footnotesize $\widehat{PV}(S, T, \epsilon, d)$ \label{PV}}
\end{algorithm}
\vspace*{-2mm}
\begin{algorithm}[H]\footnotesize
\begin{algorithmic}
\INPUT{$S\!=\!\{(\xbf_s,y_s)\}_{s=1}^{|S|}$ a source  sample, $T\!=\!\{\xbf_t\}_{t=1}^{|T|}$ a target  sample,  $\Hcal$, $\mu>0$, $\epsilon>0$,  $d$}
\OUTPUT{$\BQ(\cdot)$}
\STATE $M_{ST}$ $\leftarrow$ Step 1. and 2. $\widehat{PV}(S, T, \epsilon, d)$
\STATE $\widehat{T} \leftarrow \{(\xbf_t,y_s)\! :\!(\xbf_t,\xbf_s)\! \in\! M_{ST}, (\xbf_s,y_s)\!\in\! S\}$
\STATE return MinCq($\widehat{T},\Hcal,\mu$)
\end{algorithmic}
\caption{\footnotesize PV-MinCq($S,T,\Hcal,\mu,\epsilon,d$) \label{PV-MinCq}}
\end{algorithm}
\end{minipage}

\section{A C-bound suitable to  Domain Adaptation with Label Transfer}
\label{sec:newcbound}
First, we propose to rewrite the C-bound with a labeling function $l:X\mapsto Y$, which associates a 
label $y\in Y$ to an unlabeled example $\xbf_t\sim\DT$. 
Given such a function, the C-bound becomes:
\begin{corollary}
For all distribution $\posterior$ over  $\Hcal$, for all domain $\PT$ over $X\times Y$ of marginal (over $X$) $\DT$, for all labeling functions $l:X\mapsto Y$ such that $\esp{h\sim \posterior}\esp{\xbf_t\sim \DT} \l(\xbf_t) h(\xbf_t)> 0$, we have,\\[-0.5mm]
\centerline{$\displaystyle \RPT(\BQ) \leq 1 - \frac{\left(\esp{h\sim \posterior}\esp{\xbf_t\sim \DT} \l(\xbf_t) h(\xbf_t)\right)^2 }{\esp{(h,h')\sim \posterior^2}\esp{\xbf_t\sim \DT} h(\xbf_t)h'(\xbf_t)} +\frac{1}{2} \left|\esp{(\xbf_t,y_t)\sim \PT}\left(y_t-l(\xbf_t)\right)\right|.$}\\[-4.5mm]
%
\end{corollary}
The first two terms correspond simply to the usual C-bound measured with the labeling function $l$.
The term  
$\tfrac{1}{2}\ \left|\espdevant{(\xbf_t,y_t)\sim \PT}(y_t-l(\xbf_t))\right|$
  can be seen as a divergence between the true labeling and the one provided by $l$: The more similar $l$ and the true labeling are, the tigher the bound is.\\
With a DA point of view, an important remark is that only one domain appears in this bound. Then, we guess that this domain is the target one, and that the computation of a relevant labeling function has to make use of the information carried by the source labeled sample $S$.
Concretely, given a labeled source instance $(\xbf_s,y_s)$, we want to transfer its label $y_s$ to an unlabeled target point $\xbf_t$ close to $\xbf_s$. This will give rise to an auto-labeled target sample, on which we can apply MinCq.
To tackle the issue of defining the label transfer, we propose, in the following, to investigate a recent measure of divergence between distributions: the Perturbed Variation \cite{PV}.



\section{A Domain Adaptation MinCq with the Perturbed Variation}
\label{sec:transfer}
We first recall the definition of the Perturbed Variation (PV) proposed in \cite{PV}.
\begin{definition}[\cite{PV}]
Let $\DS$ and $\DT$ two marginal distributions over $X$, let $M(\DS,\DT)$ be the set of all joint distributions over $X\times X$ with marginals $\DS$ and $\DT$. The perturbed variation w.r.t. a distance $d:X\times X\mapsto \R$ and $\epsilon > 0$ is defined by,\\[1mm]
\centerline{$PV(\DS,\DT, \epsilon, d) = \inf_{\mu\in M(\DS,\DT)} \Prob{\mu}\left[d(\mathcal{X},\mathcal{X}')> \epsilon\right],$}
over all pairs $(\DS,\DT)\sim \mu$, such that the marginal of $\mathcal{X}$ (resp. $\mathcal{X'}$) is $\DS$ (resp. $\DT$). \\[-6.5mm]
\end{definition}
In other words, two samples are similar if every target instance is close to a source instance.
Note that this measure is consistent and that its empirical counterpart $\widehat{PV}(S,T)$ can be efficiently computed by a maximum graph matching procedure described in Algo. \ref{PV} \cite{PV}.\\
In our label transfer objective, we then propose to make use of the maximum graph matching computed $M_{ST}$ by the PV at step $2$ of  Algo. \ref{PV} (with $d$ the euclidian distance and $\epsilon$ a hyperparameter). Concretely, we label the examples from the unlabeled target sample $T$ with $M_{ST}$, with the intuition that if $\xbf_t\in T$ belongs to a pair $(\xbf_t,\xbf_s)\in M_{ST}$, then $\xbf_t$ is affected by the true label of $\xbf_s$. Else, we remove $\xbf_t$ from $T$. The auto-labeled sample obtained is denoted by $\widehat{T}$.
Then we provide   $\widehat{T}$ to MinCq. Our global procedure, called \mbox{PV-MinCq}, is summarized in Algo. \ref{PV-MinCq}.\\
Obviously, a last question concerns the hyperparameters selection.
Usually in DA, one can make use of a reverse/circular validation as done in \cite{BruzzoneM10,PBDA,DASF12}.
However, since in our specific situation  with \mbox{PV-MinCq}, we have not directly make use of the value of the PV, we propose to select parameters with a $k$-fold validation process optimizing the trade-off: $\RS(\BQ) + \widehat{PV}(S,T)$, where $\RS(\BQ)$ is the empirical risk on the source sample. 
This heuristic is justified by the philosophy of DA: Minimize the divergence (measured with the PV) between the domains while keeping good performances on the source labels transferred on the target points.

\section{Experimental Results}
\label{sec:expe}
We tackle the 
toy problem called ``inter-twinning moon'', each moon  corresponds to one class.
We consider seven target domains rotating anticlockwise the source domain according to $7$ angles.
Our \mbox{PV-MinCq} is compared with  MinCq and SVM with no adaptation, and with DA approaches: The semi-supervised Transductive-SVM (TSVM) \cite{Joachims99},  the iterative DA algorithms DASVM \cite{BruzzoneM10} (based on an auto-labeling) and DASF~\cite{DASF12} (based on the usual bound in DA \cite{BenDavid-NIPS07}), and the PAC-Bayesian DA method PBDA \cite{PBDA}.
We also report a version of MinCq that makes use of a $k$-NN based auto-labeling (NN-MinCq): We label a target point with a $k$-NN classifier of which the prototypes comes from the source sample.
We used a Gaussian kernel for all the methods. The preliminary results -- illustrated on Tab. \ref{tab:res} -- are very promising.
Firstly, \mbox{PV-MinCq} outperformns on average the others, and appears more robust to change of density (NN-MinCq and MinCq appears also more robust). 
This confirms the importance to take into account the disagreement between voters in DA\footnote{Note that preliminary experiments using PV with a SVM have implied poor results. This also probably confirms the importance of the disagreement.}.
Secondly, the PV-based labeling implies better results than the NN one.
Unlike a NN-based labeling, using the matching implied by the computation of the PV appears to be a colloquial  way to control the divergence between domains since it clearly focuses on high density region by removing the target example without matched source instance, in other words on regions where the domains are close.
These two points confirm that the PV is a relevant measure to control the process for a DA task.

\begin{table}[t]
\footnotesize
\centering
\begin{tabular}{|c||c|c|c|c|c|c|c|c|}
          \hline
          Target rotation angle & $\quad20\degree\quad$\hspace*{-0.17cm} & \hspace*{-0.17cm}$\quad30\degree\quad$\hspace*{-0.17cm} &\hspace*{-0.17cm} $\quad40\degree\quad$ \hspace*{-0.17cm}& \hspace*{-0.17cm}$\quad50\degree\quad$ \hspace*{-0.17cm} & \hspace*{-0.17cm}$\quad60\degree\quad$ \hspace*{-0.17cm}&\hspace*{-0.17cm}$\quad70\degree\quad$ \hspace*{-0.17cm}&\hspace*{-0.17cm}$\quad80\degree\quad$ \hspace*{-0.17cm}\\ 
          \hline
          \hline
          MinCq & $92.1$ & $78.2$ & 	$69.8$ &	$61$ &$ 50.1$ &$40.7$ &	$32.7$ \\
          \hline
          SVM            &  $89.6$     & $76$     &  $68.8$    &  $60$   & $47.18$  &  $26.12$    & $19.22$ \\
          \hline
          \hline
          TSVM          &   $\mathbf{100}$     &  $78.9$    &  $74.6$    &  $70.9$   & $64.72$ & $21.28$    &  $18.92$    \\ 
\hline
DASVM  & $\mathbf{100}$ & $78.4$& $71.6$&$66.6$ &$61.57$ & $25.34$ & $21.07$\\ 
\hline
PBDA & $90.6$  & $89.7$ & $77.5$ & $58.8$ & $42.4$ & $37.4$ & $39.6$ \\ 
\hline
         DASF          &  $98$ & $92$ & $83$ & $70$ & $54$ &$ 43$ &$38$ \\ 
\hline
NN-MinCq  & $97.7$ & $83.7$ & $77.7$ & $69.2$ & $58.1$ & $47.9$ & $42.1$ \\
\hline 
\hline 
PV-MinCq & $99.9$ & $\mathbf{99.7}$ & $\mathbf{99}$ & $\mathbf{91.6}$ & $\mathbf{75.3}$ & $\mathbf{66.2}$ &  $\mathbf{58.9}$ \\ 
\hline
\end{tabular}
\caption{Average accuracy results on $10$ runs for $7$ rotation angles.\label{tab:res}}
\end{table}
\section{Conclusion and Future Work}
In this work, we have proposed a first procedure to tackle DA by making use of the recent algorithm called MinCq.
Indeed, MinCq allows us to take into account the disagreement between classifiers, which is known to be crucial  in DA.
Our approach has the originality to directly minimize a risk on the target domain thanks to a labeling defined with the Perturbed Variation distance between distributions.
The preliminary results obtained are promising, and we would like to apply the method to real-life applications.
 Another exciting perspective is to define new label transfer functions, for example by computing the PV with a more adapted distance $d$ such as the domain disagreement.

\bibliography{biblio}
\bibliographystyle{unsrt}

\end{document}